\documentclass{article}
\usepackage{spconf,amsmath,graphicx}

\usepackage{graphics,graphicx} 
\usepackage{epsfig} 
\usepackage{mathptmx} 
\usepackage{times} 
\usepackage{amsmath} 
\usepackage{amssymb}  
\usepackage{url}
\usepackage{booktabs}
\usepackage{caption}
\usepackage{multirow, multicol}
\usepackage{color}
\usepackage{gensymb}

\def\ours{FAWN}
\def\L{\mathcal{L}}
\def\F{\mathcal{F}}
\def\W{\mathcal{W}}

\definecolor{unitednationsblue}{rgb}{0.36, 0.57, 0.9}
\definecolor{fuchsiapink}{rgb}{1.0, 0.47, 1.0}
\definecolor{mint}{rgb}{0.24, 0.71, 0.54}
\definecolor{fawn}{rgb}{0.9, 0.67, 0.44}
\definecolor{lightslategray}{rgb}{0.47, 0.53, 0.6}

\newcommand{\wallscolor}[1]{{\textcolor{unitednationsblue}{\textbf{\textit{#1}}}}}
\newcommand{\floorcolor}[1]{{\textcolor{mint}{\textbf{\textit{#1}}}}}
\newcommand{\othercolor}[1]{{\textcolor{fuchsiapink}{\textbf{\textit{#1}}}}}

\newcommand{\ourscolor}[1]{{\textcolor{fawn}{\textbf{#1}}}}
\newcommand{\baselinecolor}[1]{{\textcolor{lightslategray}{\textbf{#1}}}}

\def\L{{\cal L}}

\title{\LARGE \bf
FAWN: Floor-And-Walls Normal Regularization for Direct Neural TSDF Reconstruction}
\name{Anna Sokolova, Anna Vorontsova, Bulat Gabdullin, and Alexander Limonov}
\address{
        Samsung Research
    }

\begin{document}
%
\maketitle
\begin{abstract}

Leveraging 3D semantics for direct 3D reconstruction has a great potential yet unleashed. For instance, by assuming that walls are vertical, and a floor is planar and horizontal, we can correct distorted room shapes and eliminate local artifacts such as holes, pits, and hills. In this paper, we propose \ours{}, a modification of truncated signed distance function (TSDF) reconstruction methods, which considers scene structure by detecting walls and floor in a scene, and penalizing the corresponding surface normals for deviating from the horizontal and vertical directions. Implemented as a 3D sparse convolutional module, \ours{} can be incorporated into any trainable pipeline that predicts TSDF. Since \ours{} requires 3D semantics only for training, no additional limitations on further use are imposed. We demonstrate, that \ours{}-modified methods use semantics more effectively, than existing semantic-based approaches. Besides, we apply our modification to state-of-the-art TSDF reconstruction methods, and demonstrate a quality gain in \textsc{ScanNet}, \textsc{ICL-NUIM}, \textsc{TUM RGB-D}, and \textsc{7Scenes} benchmarks.

\end{abstract}
\section{Introduction}
\label{sec:intro}

\begin{figure}[t!]
\setlength{\tabcolsep}{3pt}
\begin{tabular}{cc}
    Deviation of \floorcolor{floor} normals from the vertical direction and \\
    \wallscolor{walls} normals from the horizontal direction \\
   \includegraphics[width=1\columnwidth]{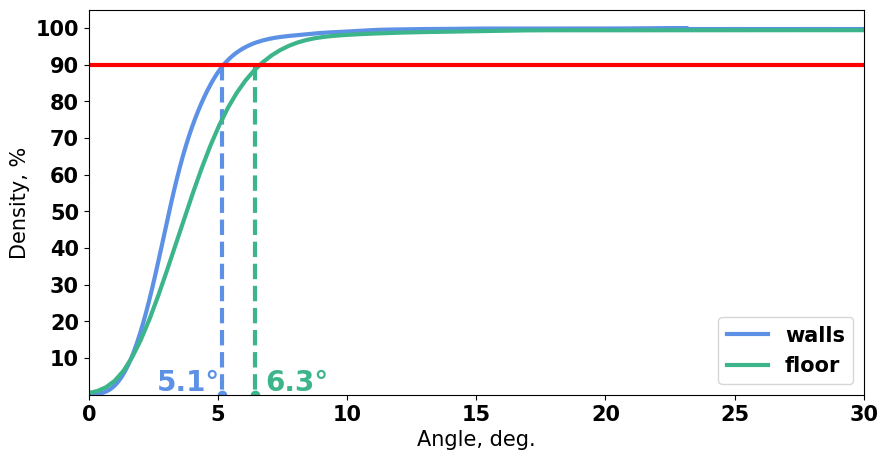}
\end{tabular}
\caption{
We verify \ours{} assumption on ScanNet. Due to imperfect ground truth scans, floor and walls normals deviate from the vertical and horizontal direction, yet the median deviation should be minor. In fact, the median angle between \floorcolor{floor} normals and the vertical direction is within $5.1\degree$, while the median angle between \wallscolor{walls} normals and the horizontal direction is within $6.3\degree$ for 90\% of scenes.}
\label{fig:fawn-assumption}
\end{figure}

3D reconstruction is a core problem in computer vision, facilitating applications such as robotics and AR/VR. These usage scenarios require fine-detailed, plausible, and dense reconstructions. Quality scans can be acquired with a precise, high-end capturing platforms. For consumer devices such as smartphones or robot vacuum cleaners, obtaining such reconstructions still remains a challenge due to imperfection of input data: incompleteness of observations in the form of occlusions and unobserved regions, between-frame inconsistency, and inevitable camera pose errors. Accordingly, reconstruction artifacts might appear, such as holes in occluded and unobserved regions or non-planar walls and floor either malformed or covered with pits and hills. While not being dramatic, such artifacts still impose limitations, e.g., for path planning for robots where room boundaries should be estimated precisely. Dense 3D reconstruction from RGB images traditionally implies estimating depth maps for RGB images and fusing them, which is another potential source of errors. Recent 3D reconstruction methods try minimizing the undesired effect with the direct prediction of TSDF. Such methods extract image features using a 2D CNN, backproject them into a 3D space, aggregate them, and predict the final TSDF volume using a 3D CNN. Being an efficient way to consider all input images jointly, using TSDF does not resolve the issues with global scene structure and planar surfaces. Fortunately, 3D semantics provides valuable clues about the scene structure, which we formulate as a \textit{\ours{} assumption: in a vast majority of indoor spaces, walls are vertical, and a floor is horizontal} (Fig.~\ref{fig:fawn-assumption}). 
This basic, non-restrictive knowledge about indoor scene geometry might be leveraged in the reconstruction process. In this paper, we propose \ours, a modification of direct reconstruction methods, which operates under a \ours{} assumption. We find walls and a floor in a scene and constrain them, penalizing wall normals for having a non-zero vertical component, and floor normals -- for deviation from the vertical direction. 

\begin{figure}[t!]
\setlength{\tabcolsep}{3pt}
\begin{tabular}{ccc}
    Ground truth scan & VoRTX\cite{stier2021vortx} & VoRTX+\ours{} \\
   \includegraphics[width=0.32\columnwidth]{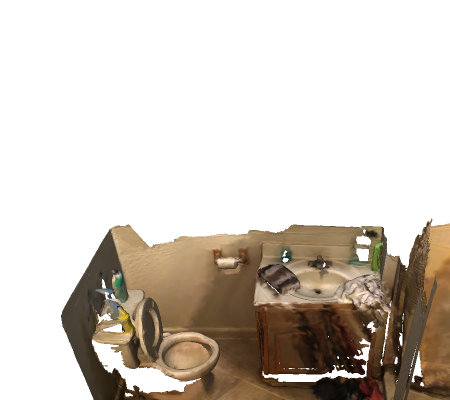} &
    \includegraphics[width=0.32\columnwidth]{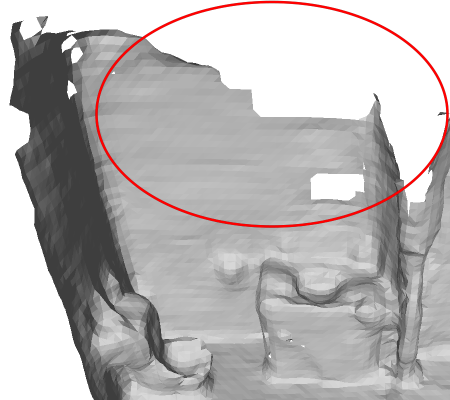} &
    \includegraphics[width=0.32\columnwidth]{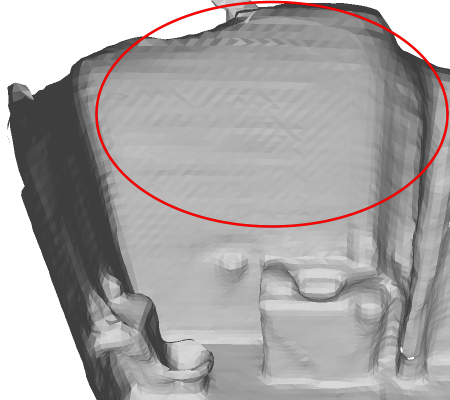} \\
\end{tabular}
\caption{3D reconstructions by state-of-the-art VoRTX~\cite{stier2021vortx} and VoRTX+\ours{}. \ours{} modification improves reconstruction of planar regions and fills the holes even in the areas not covered by ground truth.}
\label{fig:qualitative-comparison-vortx-with-gt}
\end{figure}

Our main contributions are as follows:
\begin{itemize}
\item We present \ours{}, a modification of 3D scene reconstruction methods, which includes a novel trainable module and an associated training procedure guided with normal-semantic losses. \ours{} can be added to an arbitrary trainable pipeline that outputs TSDF, and trained end-to-end on semantically annotated 3D data;
\item We apply \ours{} to several state-of-the-art reconstruction approaches and demonstrate a performance gain over the state-of-the-art on multiple datasets.
\item We discuss the standard evaluation protocol, and propose extending it with a novel \textit{coverage} score measuring the completeness of a reconstructed scan: such a fundamental feature has not been explored previously and hence is not represented in existing quality metrics.
\end{itemize}

\section{Related Work}
\label{sec:related}

\subsection{TSDF Prediction}

TSDF is an efficient way to consider all images of a scene jointly, so 3D scene reconstruction can be solved by predicting TSDF volume. The progress of TSDF reconstruction approaches is mainly associated with elaborate feature aggregation strategies. The first end-to-end 3D reconstruction approach, Atlas~\cite{murez2020atlas}, back-projects features into a voxel volume, fuses features within each voxel by simple averaging, and finally predicts a TSDF volume with a 3D CNN. NeuralRecon~\cite{sun2021neuralrecon} runs a hierarchical fusion strategy on sequential frames, averaging features of nearby views and fusing across clusters with an RNN network. VisFusion~\cite{gao2023visfusion} also fuses features with an RNN, but takes visibility into account and sparsifies voxels to improve efficiency. On the contrary, VoRTX~\cite{stier2021vortx} fuses all views jointly using a transformer model. Another transformer-based model, TransformerFusion~\cite{bozic2021transformerfusion}, attends to the image features, and uses the attention maps to select frames during the inference.

\subsection{Using Normals for 3D Reconstruction}

Using surface normals for 3D reconstruction has been actively investigated over the past years. VolSDF~\cite{yariv2021volume} and NeUS~\cite{wang2021neus} minimize photometric loss and constraint SDF with the eikonal loss, encouraging SDF normals to have a norm of 1. NeurIS~\cite{wang2022neuris} regularizes rendered normals with the normals predicted by a trainable method, and filters constraints by multi-view photometric consistency between normals and depths. NeuralRoom~\cite{wang2022neuralroom} also uses a normal estimation network, and applies normal loss for textureless regions, where photometric loss fails due to shape-radiance ambiguity. Unlike NeurIS and NeuralRoom, we acquire normals directly from the predicted TSDF. Moreover, all the models mentioned are implicit: they are trained for each given scene, hence hardly scalable. Models with \ours{}, once trained, can be simply and quickly inferred on any scene.

\begin{figure*}[ht!]
    \centerline{\includegraphics[width=0.95\textwidth]{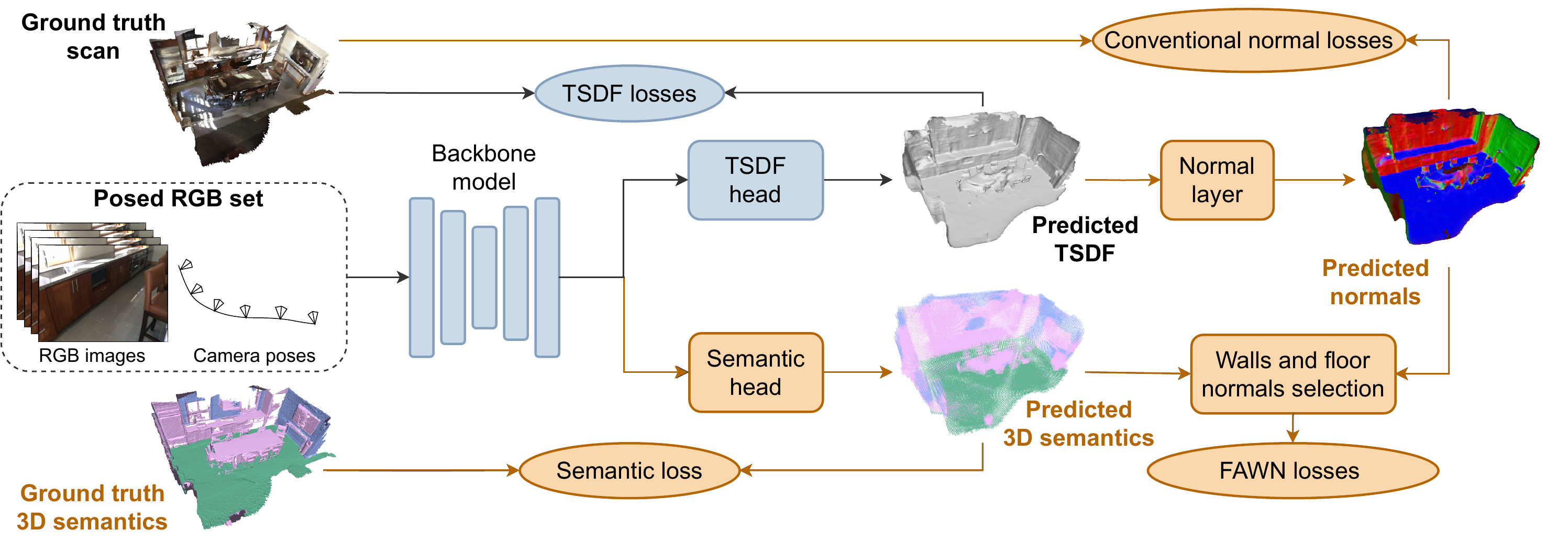}}
    \caption{Training procedure of a \ours-modified method, with \baselinecolor{baseline components of the pipeline} and \ourscolor{\ours{} add-ons} specified. A set of RGB images with camera poses are first processed with a backbone. The extracted features are passed to a baseline TSDF head and \ours{} auxiliary semantic head, that detects walls and floor. The semantic head is guided with a \textit{semantic loss} $\L_{sem}$ during the training, while the TSDF head is trained to minimize baseline-specific \textit{TSDF losses} $\L_{TSDF}$. Surface normals are derived from the predicted TSDF as first-order gradients, and are used to regularize geometry via normal losses $\L_{norm}$. Our key novelty is in combining normals and semantics in $\L_{FAWN}$: surface normals in \wallscolor{walls} regions are penalized for deviation from the horizontal direction, while the constraints imposed on \floorcolor{floor} normals force them to be vertical. As a result, the floor and walls in a reconstructed scene become more smooth and planar, and holes get filled with planar segments.}
\end{figure*}

\subsection{Using Semantics for 3D Reconstruction}

Various ways of utilizing semantics for 3D reconstruction have also been considered. SceneCode~\cite{zhi2019scenecode} obtains a semantic representation with VAE conditioned on an RGB image, and fuses semantic labels by jointly optimizing latent codes of overlapping images. Numerous methods~\cite{avetisyan2019endtoend, avetisyan2019scan2cad, dahnert2019embedding, hampali2021montecarlo, ainetter2023automatically} recognize objects and replace them with CAD models: while providing visually plausible results, this is hardly restoring a real scene, but creating a 3D model more or less resembling it. On the contrary, we aim to reconstruct the actual scene. 
Implicit Manhattan-SDF~\cite{guo2022manhattan} also combines surface normals and semantics for 3D reconstruction, forcing floor and walls normals to be collinear with three dominant directions. Compared to the Manhattan assumption, \ours{} assumption is less restrictive and relevant for a wider range of real scenes. Semantic information is given as 2D semantic maps, predicted with an off-the-shelf DeepLabV3+ model. On the contrary, \ours{} does not rely on auxiliary models and is trained end-to-end to predict 3D semantics directly. Atlas~\cite{murez2020atlas} fuses features into a single feature volume and also has a semantic head, but it actually aims at identifying semantics, while \ours{} treats semantics as an additional source of spatial information; \textit{a guidance} rather than \textit{a target}.


\section{Proposed Method}
\label{sec:method}

Our key contribution is aligning normals for floor and walls surfaces. Prior to the alignment, we need to segment a floor and walls in a scene; to this end, we train a semantic head that addresses a 3D semantic segmentation task. Below, we describe this semantic head, its training protocol, and floor-and-walls-normal (\ours) losses.

\subsection{Auxiliary Semantic Head}

The semantic head is a decoder part of a 3D sparse U-Net-like~\cite{unet} CNN, that branches off from the backbone. It consists of two 3D sparse convolutional modules. Each module comprises a sparse 3D convolution, followed by batch normalization and two residual 3D convolutional blocks. Our semantic head outputs 3D semantics in a form of a spatial map with three channels, each corresponding to a single semantic category: \wallscolor{walls}, \floorcolor{floor}, and \othercolor{other}. 

\subsection{Training}

We train the semantic head on \textsc{ScanNet} point clouds annotated with per-point semantic labels. Original \textsc{ScanNet} categories \textit{floor} and \textit{carpet} are remapped into the \textit{floor} category, while \textit{wall}, \textit{window}, \textit{door}, \textit{picture}, and \textit{whiteboard} comprise the \textit{walls} category. All points, considered belonging neither to \textit{walls} nor \textit{floor} categories, fell under the \textit{other} category. We do not consider a separate \textit{ceiling} category, since there are too few ceiling points in \textsc{ScanNet} scans due to the capturing process; yet we assume our method can be easily extended to handle ceiling in the same way as floor.

The training is performed in two stages. At the first stage, the main model that predicts TSDF is trained jointly with the semantic head, that learns to divide points into three categories guided by semantic losses. We do not penalize scene geometry during the first epochs so that the entire training procedure does not get disrupted by early erroneous estimates. In the second stage, we add losses on surface normals and proceed with a joint prediction of the scene geometry and semantics. Surface normals are derived from an estimated TSDF as the first-order gradients. Just as image gradients are computed in the 2D case~\cite{sobel} with a 2D convolution, we implement normal estimation via a frozen 3D convolution, which can be easily incorporated into a trainable pipeline. Note, that it does not cause a computational overhead, since the scene surface is not derived during the optimization.

\subsection{Losses}

For each baseline, we apply TSDF losses $\L_{TSDF}$ used in the original training procedures. Besides, wrong semantic predictions are penalized with the standard cross-entropy loss $\L_{sem}$.

Additionally, we use losses based on surface normals in the second stage of training. First, we introduce floor-and-walls-normal, or \ours, regularization that considers both surface normals and 3D semantics. Generally speaking, we encourage walls to be vertical, and a floor to be horizontal. 
\begin{figure*}[ht!]
\begin{tabular}{ccccc}
    \includegraphics[width=0.175\linewidth]{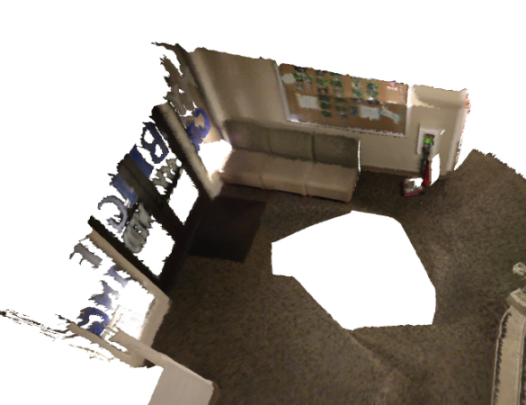} & 
    \includegraphics[width=0.175\linewidth]{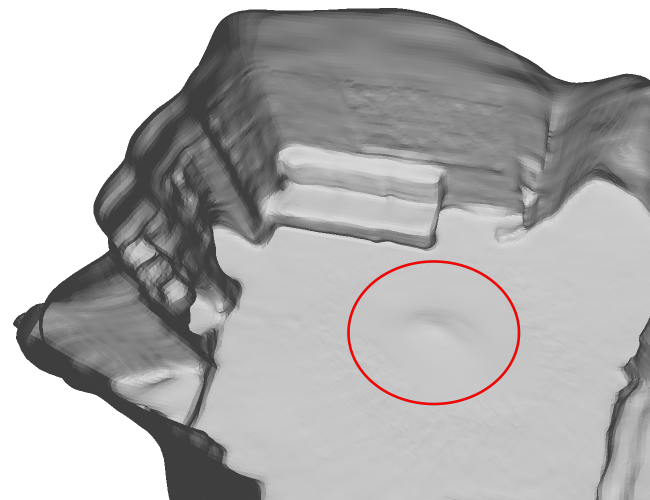} &
    \includegraphics[width=0.175\linewidth]{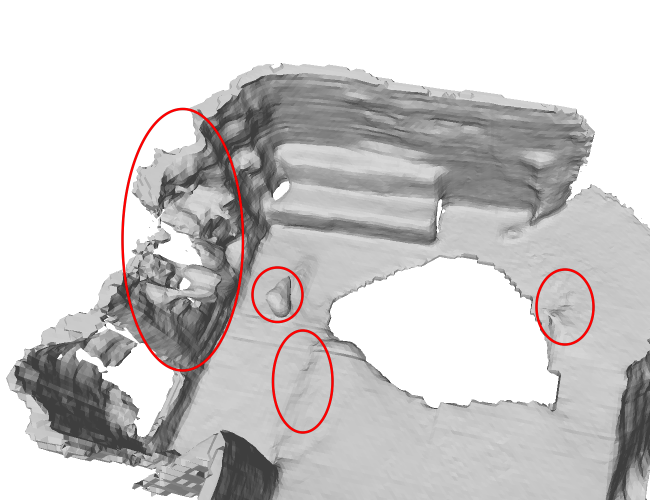} &
    \includegraphics[width=0.175\linewidth]{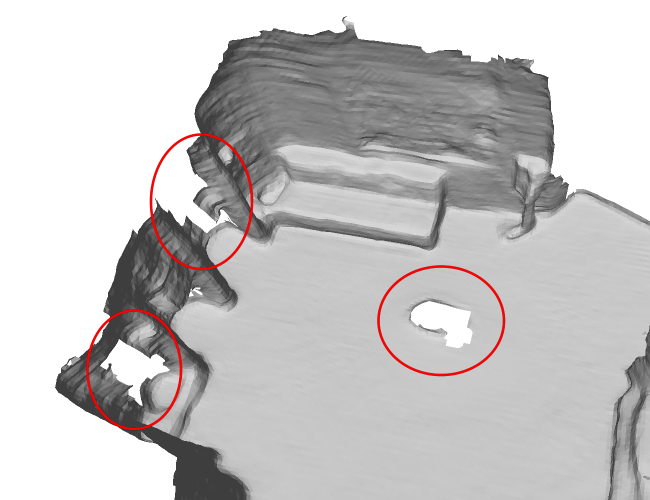} &
    \includegraphics[width=0.175\linewidth]{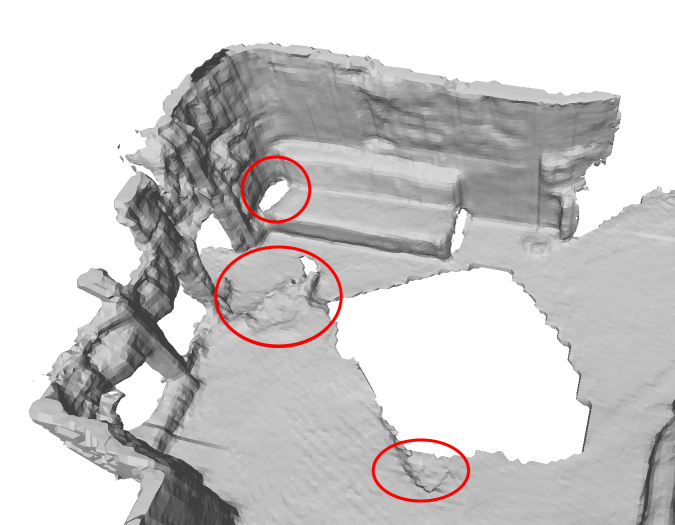} \\
    Ground truth scan & Atlas~\cite{murez2020atlas} & NeuralRecon~\cite{sun2021neuralrecon} & VoRTX~\cite{stier2021vortx} & VisFusion~\cite{gao2023visfusion} \\ \\
    \includegraphics[width=0.175\linewidth]{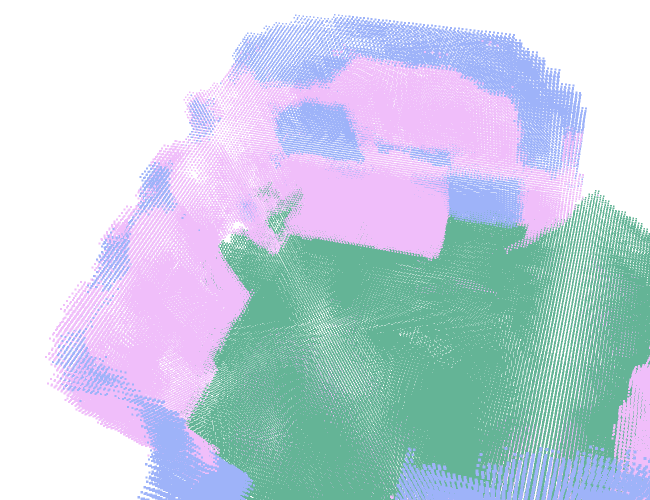} &
    \includegraphics[width=0.175\linewidth]{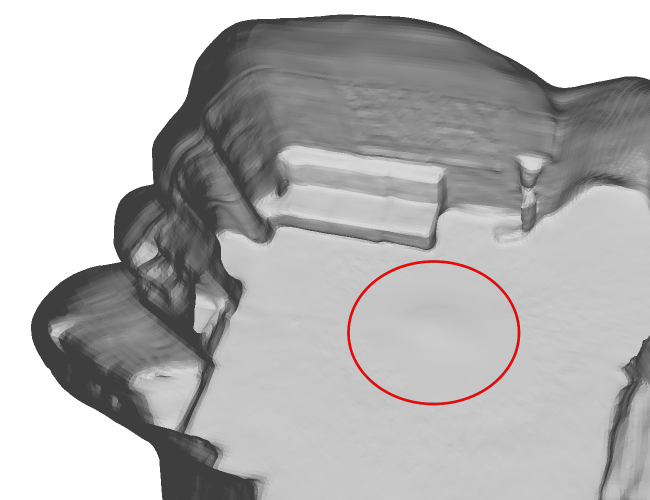} &
    \includegraphics[width=0.175\linewidth]{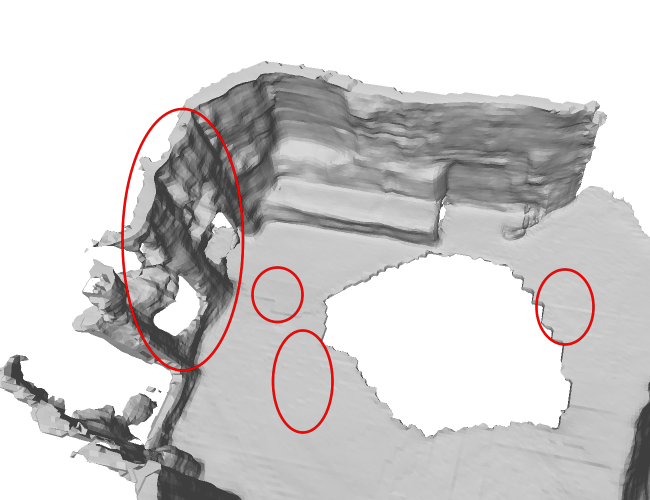} &
    \includegraphics[width=0.175\linewidth]{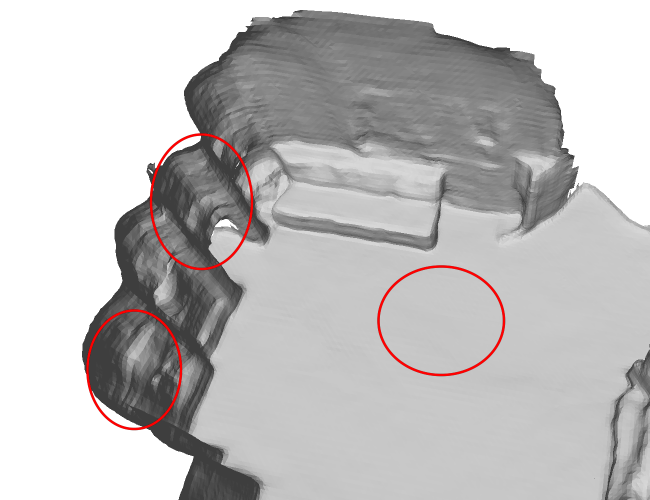} &
    \includegraphics[width=0.175\linewidth]{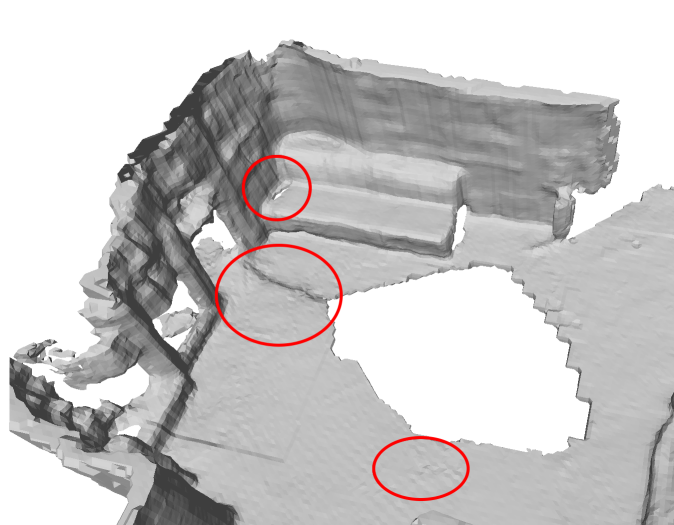} \\
    Predicted 3D semantics & Atlas+\ours{} & NeuralRecon+\ours{} & VoRTX+\ours{} & VisFusion+\ours{} \\
    (\wallscolor{walls}, \floorcolor{floor}, \othercolor{other}) & & & & \\
\end{tabular}
\caption{3D reconstructions of a \textsc{ScanNet} scene obtained with baseline methods and their \ours-modified versions.}
\label{fig:qualitative-comparison-scannet}
\end{figure*}

\begin{table*}[h!]
\centering
\caption{Reconstruction quality \ours{} on \textsc{ScanNet} and \textsc{TUM RGB-D}. The best scores are \textbf{bold}.}
\label{tab:results-scannet-tum}
\setlength{\tabcolsep}{3pt}
\begin{tabular}{lcccccccccccc}
\toprule
\multirow{3}{*}{Method} & \multicolumn{6}{c}{ScanNet} & \multicolumn{6}{c}{TUM RGB-D} \\
\cmidrule{2-7} \cmidrule{8-13}
& Acc$\downarrow$ & Comp$\downarrow$ & Prec$\uparrow$ & Rec$\uparrow$ & F-score$\uparrow$ & Cover$\uparrow$ & Acc$\downarrow$ & Comp$\downarrow$ & Prec$\uparrow$ & Rec$\uparrow$ & F-score$\uparrow$ & Cover$\uparrow$\\
& [cm] & [cm] & [\%] & [\%] & [\%] & [\%] & [cm] & [cm] & [\%] & [\%] & [\%] & [\%] \\ 
\midrule
Ground truth & - & - & - & - & - & 94.6 & - & - & - & - & - & 72.8 \\
\midrule
Atlas~\cite{murez2020atlas}                 & 6.8 & 9.8 & 64.0 & 53.9 & 58.3 & \textbf{99.9} & 19.7 & 6.6 & 43.9 & 56.9 & 48.4 & 99.9 \\
Atlas + \ours{}         & \textbf{6.3} & \textbf{8.5} & \textbf{67.0} & \textbf{57.8} & \textbf{61.9} & \textbf{99.9} & \textbf{17.3} & \textbf{6.3} & \textbf{47.0} & \textbf{58.7} & \textbf{50.9} & \textbf{100.0} \\
\midrule
NeuralRecon~\cite{sun2021neuralrecon}           & \textbf{4.9} & 13.3 & 69.1 & 46.1 & 55.1 & 89.9 & 9.5 & 10.0 & 45.5 & 35.6 & 39.8 & 76.0 \\ 
NeuralRecon + \ours{}   &  \textbf{4.9} &  \textbf{12.6} & \textbf{70.3}   &  \textbf{49.2} &  \textbf{57.7} & \textbf{90.9} &  \textbf{7.6} &  \textbf{9.9} & \textbf{48.2}  & \textbf{ 39.5} &  \textbf{43.1} & \textbf{81.0} \\
\midrule
VoRTX~\cite{stier2021vortx}                 & 5.4 & 9.0 & 70.8 & 58.8 & 64.1 & 96.2 & {19.2} & \textbf{5.3} & 48.5 & 65.1 & 54.3 & 91.3 \\
VoRTX + \ours{}         & \textbf{5.0} & \textbf{8.2} & \textbf{72.7} & \textbf{61.7} & \textbf{66.6} & \textbf{96.4} & \textbf{19.0} & \textbf{5.3} & \textbf{49.5} & \textbf{65.8} & \textbf{55.1} & \textbf{93.0} \\
\midrule
VisFusion~\cite{gao2023visfusion}	              & 5.0& 10.0& 70.3& 53.9& 60.8& 92.4 &\textbf{18.6} &7.1 & 43.5& 48.0& 45.0& 86.1 \\
VisFusion + \ours{}        &\textbf{4.8} &\textbf{9.6} &\textbf{70.9} &\textbf{54.3} &\textbf{61.4} & \textbf{94.7} & 18.7& \textbf{6.6}& \textbf{48.4}& \textbf{54.4}& \textbf{50.4}& \textbf{87.9} \\
\bottomrule
\end{tabular}
\end{table*}

Let us denote a set of 3D points classified as \textit{floor} as $\F$, and \textit{walls} as $\W$. We project each floor normal \((n_x,n_y,n_z)\) onto the horizontal plane, getting \((n_x,n_y)\) and penalize the length of such projections:
\begin{equation}
    \L_{floor} = \dfrac{1}{|\F|} \sum \limits_{p \in \F} \| (n_x(p), n_y(p)) \|_2.
\end{equation}
Similarly, the wall normals are projected onto the vertical axis, and the length of this projection \(|n_z|\) is penalized:
\begin{equation}
\L_{walls} = \dfrac{1}{|\W|} \sum \limits_{p \in \W} | n_z(p) |,
\end{equation}
eventually, our floor-and-walls-normal loss is formulated as:
\begin{equation}
\L_{FAWN} = \L_{walls} + \L_{floor}.
\end{equation}
Besides, we exploit conventional normal losses. We extract ground truth normals from a ground truth TSDF representation and penalize the divergence between predicted and ground truth normals using both cosine and Euclidean distances, as in~\cite{yu2022monosdf}. These reference-based losses are complemented with a non-reference eikonal loss~\cite{icml2020_2086}. We use a sum of three losses for training, denoted as $\L_{norm}$ for brevity. 

Overall, we aim to minimize the following total loss:
\begin{equation}
\L = \L_{TSDF} + \lambda_{sem}\L_{sem} + \lambda_{norm}\L_{norm} + \lambda_{FAWN}\L_{FAWN},
\end{equation}
where $\lambda_{sem}$, $\lambda_{norm}$, and $\lambda_{FAWN}$ denote weights of loss terms.


\subsection{Inference}

The trained model may be inferred on any indoor scene, no matter the presence of a floor and walls, as no extra assumptions are made on the scene structure. Moreover, neither ground truth nor predicted 3D semantics is required: 3D semantics serves for training only as an extra guidance for a TSDF-predicting model. Overall, \ours{} does not impose any additional limitations on a usage scenario.

\section{Experiments}
\label{sec:experiments}

\begin{table*}[h!]
\centering
\caption{Reconstruction quality \ours{} on \textsc{ICL-NUIM} and \textsc{7Scenes}. The best scores are \textbf{bold}.}
\label{tab:results-iclnuim-7scenes}
\setlength{\tabcolsep}{3pt}
\begin{tabular}{lcccccccccccc}
\toprule
\multirow{3}{*}{Method} & \multicolumn{6}{c}{ICL-NUIM} & \multicolumn{6}{c}{7Scenes} \\
\cmidrule{2-7} \cmidrule{8-13}
& Acc$\downarrow$ & Comp$\downarrow$ & Prec$\uparrow$ & Recall$\uparrow$ & F-score$\uparrow$ & Cover$\uparrow$ & Acc$\downarrow$ & Comp$\downarrow$ & Prec$\uparrow$ & Recall$\uparrow$ & F-score$\uparrow$ & Cover$\uparrow$\\
& [cm] & [cm] & [\%] & [\%] & [\%] & [\%] & [cm] & [cm] & [\%] & [\%] & [\%] & [\%] \\ 
\midrule
Ground truth & - & - & - & - & - & 96.1 & - & - & - & - & - & 97.8 \\
\midrule
Atlas~\cite{murez2020atlas}    & 17.5 & \textbf{31.4} & \textbf{28.0} & 19.4 & 22.9 & \textbf{100.0} & 13.9 & \textbf{27.8} & 32.0 & 24.2 & 27.2 & \textbf{99.7} \\
Atlas + \ours{}         & \textbf{17.4} & 37.7 & \textbf{28.0} & \textbf{20.2} & \textbf{23.4} & \textbf{100.0} & \textbf{12.9} & 28.4 & \textbf{33.8} & \textbf{24.9} & \textbf{28.4} & \textbf{99.7} \\
\midrule
NeuralRecon~\cite{sun2021neuralrecon}           & 21.5 & 103.1 & 21.4 & 3.6 & 5.8 & 68.5 & 10.6 & 60.3 & 36.5 & 11.4 & 16.7 & 66.2 \\
NeuralRecon + \ours{}   & \textbf{15.5}  & \textbf{48.7}  & \textbf{31.6}  & \textbf{13.7} & \textbf{19.1} & \textbf{71.9}&  \textbf{10.5} &  \textbf{48.3} & \textbf{37.5}  & \textbf{15.1}  & \textbf{20.9} & \textbf{78.5}  \\
\midrule
VoRTX~\cite{stier2021vortx}    & 10.2 & \textbf{14.6} & 44.9 & \textbf{37.5} & 40.8 & 92.2 & 11.8 & 29.8 & 33.1 & 22.7 & 26.6 & 93.3 \\
VoRTX + \ours{}         & \textbf{9.8} & {22.3} & \textbf{48.2} & 36.9 & \textbf{41.7} & \textbf{92.5} & \textbf{11.7} & \textbf{29.3} & \textbf{33.6} & \textbf{23.1} & \textbf{27.1} & \textbf{96.4} \\
\midrule
VisFusion~\cite{gao2023visfusion}	              &25.1 &34.9 & 19.5&20.0 & 20.3 & 74.7 &15.7 &39.5 &27.6 &16.1 &20.0 & 75.7 \\
VisFusion + \ours{}        &\textbf{14.7} &\textbf{29.6} &\textbf{27.1} &\textbf{33.8} &\textbf{29.9} & \textbf{79.5} & \textbf{13.9}&\textbf{39.0} &\textbf{31.5} & \textbf{18.0}& \textbf{22.5}& \textbf{78.5} \\
\bottomrule
\end{tabular}
\end{table*}

\subsection{Baselines}

First, we compare \ours{} with a way of using semantics proposed in \textbf{Atlas~\cite{murez2020atlas}}. We do not modify the original Atlas semantic head, but train it with our training protocol and losses. 

Besides, we incorporate \ours{} into several state-of-the-art open-source reproducible TSDF reconstruction methods to prove it to be beneficial for various fusion approaches. RNN-based \textbf{NeuralRecon~\cite{sun2021neuralrecon}} directly reconstructs local surfaces as multiple sparse TSDFs, and fuses them with an RNN module. Transformer-based \textbf{VoRTX~\cite{stier2021vortx}} tackles multi-view feature fusion by retaining fine details by camera-conditioned fusion and handling occlusions by estimating an initial scene geometry.
State-of-the-art online \textbf{VisFusion~\cite{gao2023visfusion}} fuses features locally by trainable visibility-aware aggregation, and then globally via recurrent units, and achieves real-time performance by ray-based voxel sparsification.

\subsection{Datasets}

We train our method on \textsc{ScanNet}~\cite{dai2017scannet}, which contains 1613 indoor scenes with ground-truth camera poses, 3D reconstructions, and semantic segmentation labels. Overall, there are 2.5M RGB-D frames acquired in 707 distinct spaces. We adopt the standard splits and report results for the test subset. We also evaluate on \textsc{TUM RGB-D}~\cite{sturm2012tumrgbd}, a longstanding RGB-D SLAM benchmark.
Finally, we report the results for \textsc{ICL-NUIM}~\cite{handa2014iclnuim}, a small-scale RGB-D reconstruction benchmark with eight scenes rendered in a synthetic framework, and 7-Scenes~\cite{shotton2013scenes}, a small yet challenging indoor RGB-D dataset comprising 7 real-world indoor spaces.

\subsection{Evaluation Protocol}

\begin{figure*}[h!]
\begin{tabular}{ccccc}
    \includegraphics[width=0.175\linewidth]{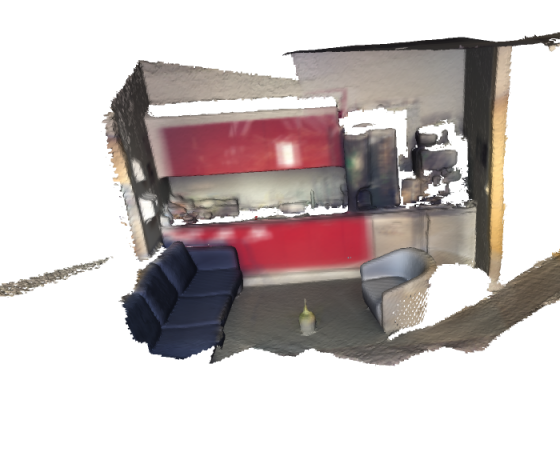} & 
    \includegraphics[width=0.175\linewidth]{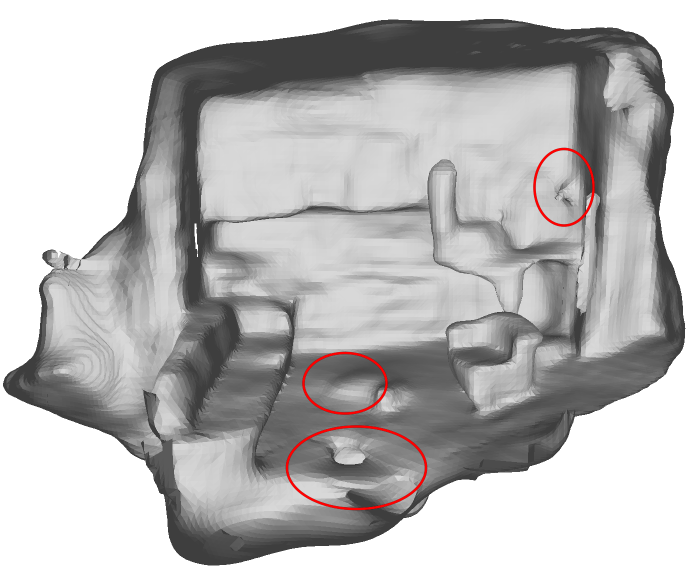} &
    \includegraphics[width=0.175\linewidth]{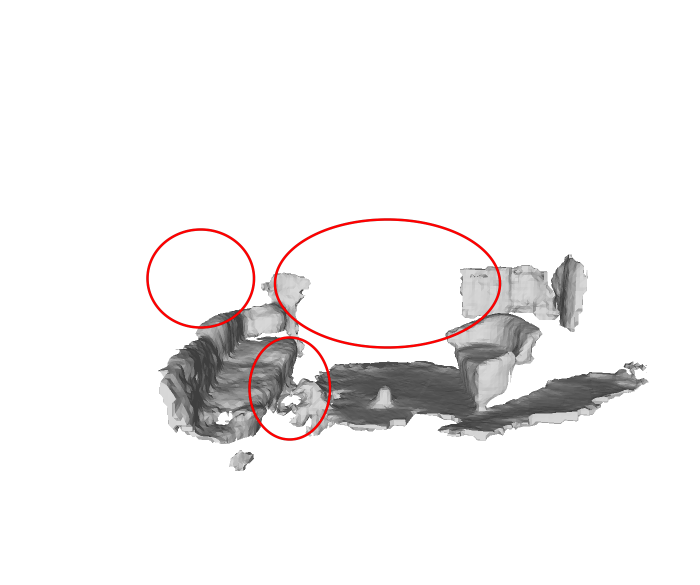} &
    \includegraphics[width=0.175\linewidth]{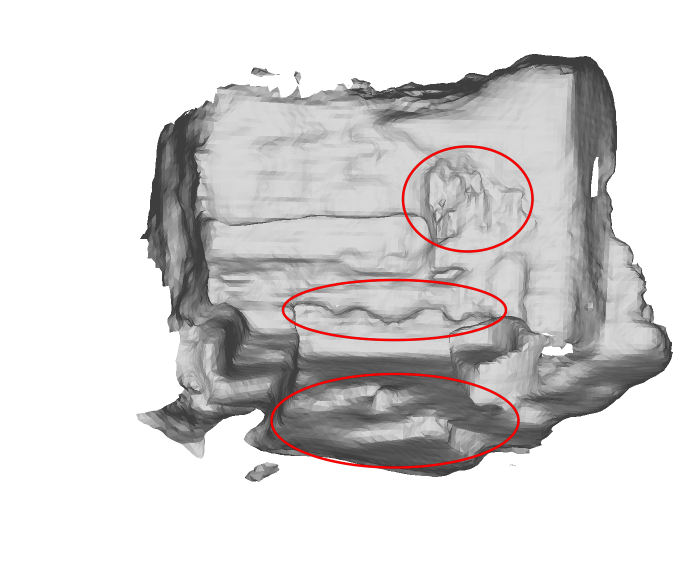} &
    \includegraphics[width=0.175\linewidth]{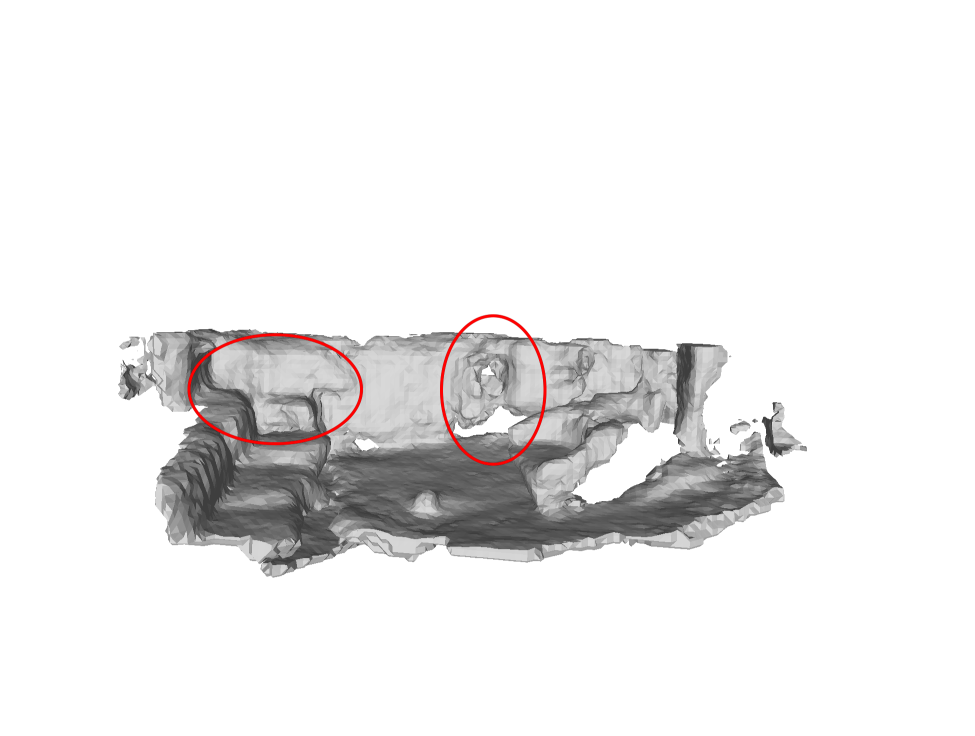} \\
     Ground truth scan & Atlas~\cite{murez2020atlas} & NeuralRecon~\cite{sun2021neuralrecon} & VoRTX~\cite{stier2021vortx} & VisFusion~\cite{gao2023visfusion} \\ 
     \textit{Coverage:} 97.2 & \textit{Coverage:} 100.0 & \textit{Coverage:} 29.0 & \textit{Coverage:} 96.0 & \textit{Coverage:} 56.8 \\ \\
    \includegraphics[width=0.175\linewidth]{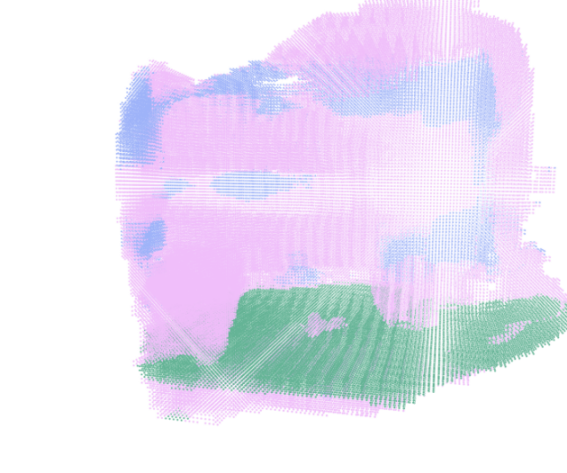} &
    \includegraphics[width=0.175\linewidth]{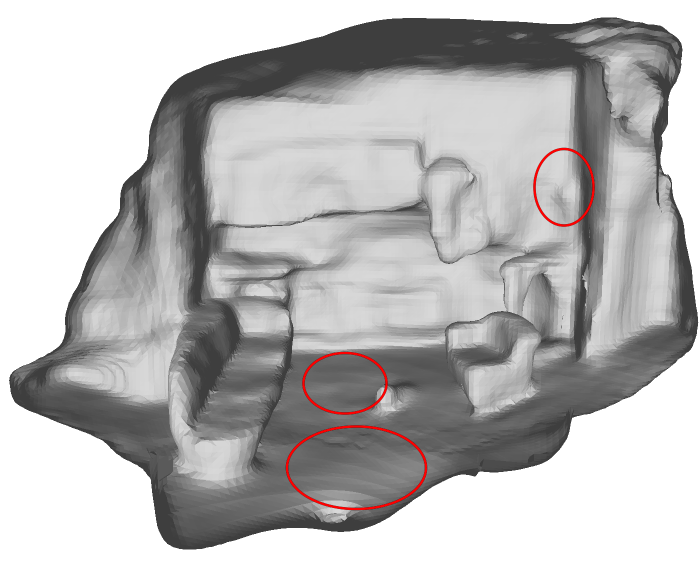} &
    \includegraphics[width=0.175\linewidth]{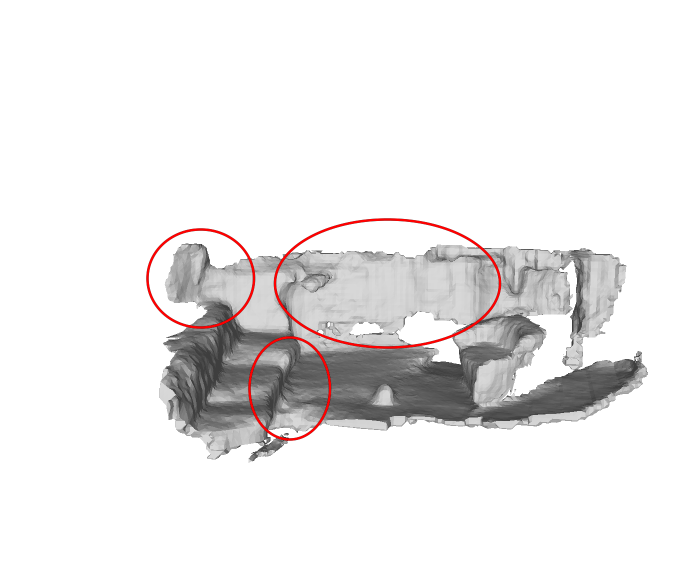} &
    \includegraphics[width=0.175\linewidth]{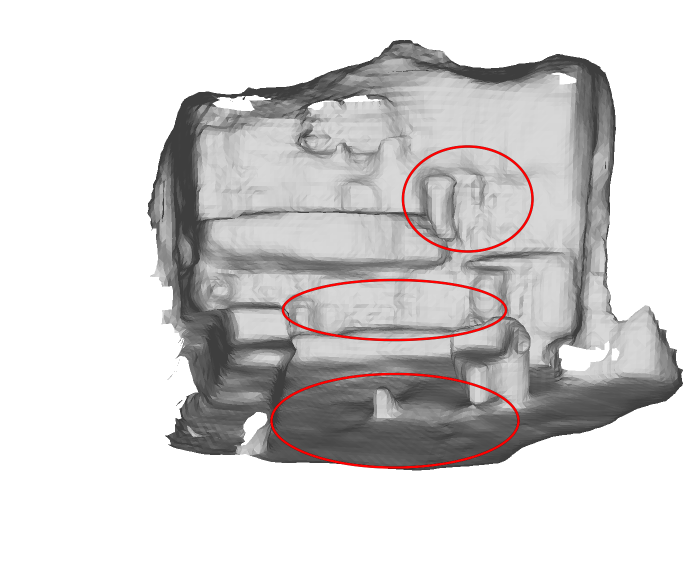} &
    \includegraphics[width=0.175\linewidth]{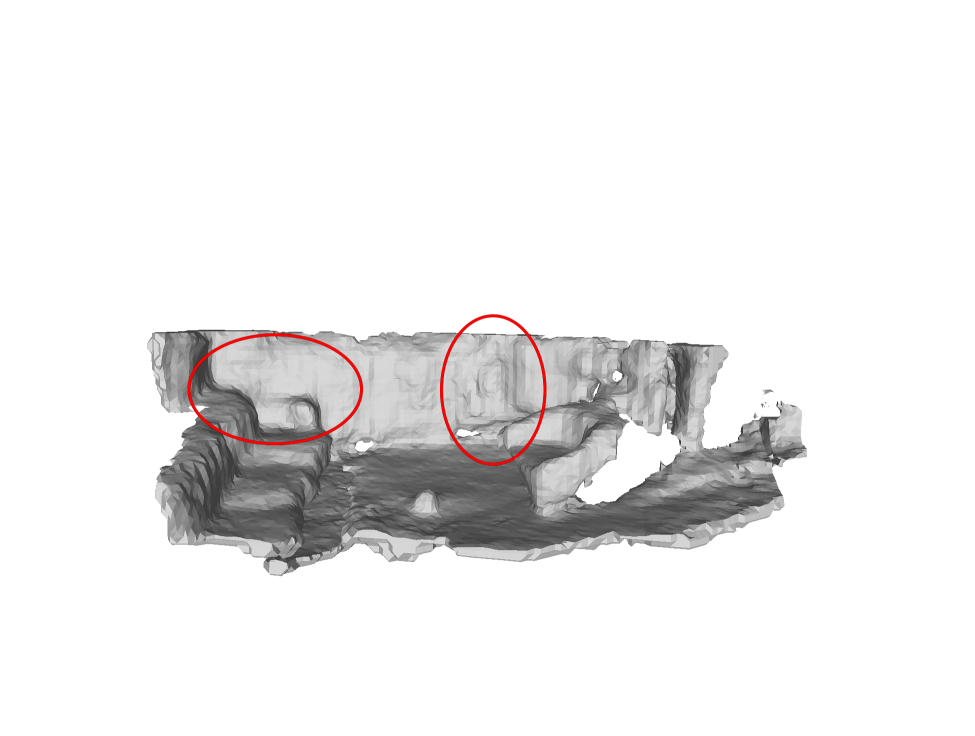} \\
    Predicted 3D semantics & Atlas+\ours{}: & NeuralRecon+\ours{} & VoRTX+\ours{} & VisFusion+\ours{}\\
    (\wallscolor{walls}, \floorcolor{floor}, \othercolor{other}) & \textit{Coverage:} 100.0 & \textit{Coverage:} 47.6 & \textit{Coverage:} 96.6 & \textit{Coverage:} 59.7 \\
\end{tabular}
\caption{3D scans of a \textsc{7Scenes} scene reconstructed with and without \ours. We report coverage to provide visual intuition: evidently, the higher the coverage score, the fewer and smaller are gaps in the reconstructed scans. The lowest score of 29.0 is obtained with the original NeuralRecon, that reconstructs only a small part of a scan. Atlas yields scans of a full coverage, but too coarse and over-smoothed, which is reflected in values of other metrics. VisFusion has a coverage of 56.8, and the scan is obviously incomplete. VoRTX provides an almost full coverage with the best reconstruction quality.} 
\label{fig:qualitative-comparison-7scenes}
\end{figure*}

We follow VoRTX~\cite{stier2021vortx} evaluation protocol for quantitative comparison. Ground truth depth maps are rendered from the ground truth mesh w.r.t. the corresponding camera poses. Similarly, estimated depth maps are rendered from the estimated mesh, masked in regions where ground truth depth is invalid, and integrated into a predicted TSDF volume. The reconstruction quality is assessed with the standard reference-based metrics~\cite{murez2020atlas} by comparing the mesh obtained from masked estimated depths with the ground truth mesh.

Yet, by applying masks, we ignore the scene parts missing in a ground truth scan. Evantually, our key benefits are not reflected in metric values. So, to highlight the superior completeness of the reconstructed scene, we formulate the novel \textit{coverage} score. To measure coverage, we render the reconstructed scene w.r.t. camera poses, frame by frame, and report an average percentage of valid pixels in such rendered images. Higher coverage values indicate a more complete reconstruction; nevertheless, coverage score should not be used solely, but combined with the reconstruction quality metrics.

\subsection{Implementation Details}

Our semantic head is trained from scratch jointly with the baseline model. We set $\lambda_{sem}=10^{-1}$, $\lambda_{norm}=10^{-4}$, $\lambda_{FAWN}=10^{-3}$. Other training settings are unchanged. All experiments are conducted using a single NVIDIA Tesla P40 GPU.

\section{Results}
\label{sec:results}

\subsection{Comparison to Prior Work}

The reconstruction metrics on \textsc{ScanNet}, \textsc{TUM RGB-D}, \textsc{ICL-NUIM}, and \textsc{7Scenes} benchmarks are gathered in Table~\ref{tab:results-scannet-tum} and Table~\ref{tab:results-iclnuim-7scenes}. \ours{} consistently improves the reconstruction quality and completeness of all the baseline approaches on all the benchmarks. Atlas is designed to ensure high completeness, so the coverage scores of the original method are already close to 100. However, as we mention before, coverage is complementary to the standard reconstruction quality measures and should not be used separately but in combination with other metrics.  

\begin{table}[h!]
\caption{Reconstruction quality obtained with various loss terms on \textsc{ScanNet}. We use $\L_{FAWN}+\L_{sem}+\L_{norm}$ as the default loss in \ours.
}
\label{tab:ablation-losses}
\setlength{\tabcolsep}{3pt}
\begin{tabular}{lccccc}
\toprule
\multirow{2}{*}{Method} & Acc$\downarrow$ & Comp$\downarrow$ & Prec$\uparrow$ & Rec$\uparrow$ & F-score$\uparrow$ \\
& [cm] & [cm] & [\%] & [\%] & [\%] \\
\midrule
VoRTX~\cite{stier2021vortx}                  & 5.4 & 9.0 & 70.8 & 58.8 & 64.1 \\
\hspace*{0.5em}$+\L_{sem}$  & 5.1 & 9.1 & 71.0 & 58.7 & 64.1 \\
\hspace*{0.5em}$+\L_{norm}$  & 5.2 & 9.2 & 71.3 & 58.8 & 64.3 \\ 
\hspace*{0.5em}$+\L_{FAWN}$  & 5.1 & 8.6 & 70.7 & 59.3 & 64.4 \\
\hspace*{0.5em}$+\L_{FAWN}+\L_{sem}$ & \textbf{5.0} & {8.5} & 72.6 & 61.1 & 66.2 \\
\hspace*{0.5em}$+\L_{FAWN}+\L_{sem}+$  & \multirow{2}{*}{\textbf{5.0}} &  \multirow{2}{*}{\textbf{8.2}} &  \multirow{2}{*}{\textbf{72.7}} &  \multirow{2}{*}{\textbf{61.7}} &  \multirow{2}{*}{\textbf{66.6}} \\
\hspace*{44pt}$+\L_{norm}$ & & & & & \\
\bottomrule
\end{tabular}
\end{table}

The scenes from \textsc{ScanNet} and \textsc{7Scenes}, reconstructed by baseline and modified approaches are depicted in Fig.~\ref{fig:qualitative-comparison-scannet} and Fig.~\ref{fig:qualitative-comparison-7scenes}. We visualize 3D semantics to mark \floorcolor{floor} and \wallscolor{walls} areas where \ours{} regularization is applied. Opposite to baselines, \ours{} encourages walls and floor to be smooth and planar, resulting in more accurate reconstructions.



\subsection{Ablation Studies}
\label{ssec:ablation}

We analyze the effect of each loss term on the reconstruction quality. We use VoRTX, since the performance gain is the most evident, so the effect of \ours{} can be studied in details. 

As can be observed in Table~\ref{tab:ablation-losses}, just a joint training of TSDF and semantics ($\L_{sem}$) does not have a significant effect on reconstruction quality. This is expected, since the tasks are basically not related. \ours{} establishes a correspondence between normals and semantics ($\L_{FAWN}+\L_{sem}$), that results in a notable improvement. $\L_{norm}$ solely has a minor effect on performance, yet brings the most significant gain when combined with $\L_{FAWN}$ and $\L_{sem}$. Accordingly, we use the most beneficial combination of $\L_{FAWN}+\L_{sem}+\L_{norm}$ by default.

An interesting outcome is that \ours{} with a trained semantic head ($\L_{FAWN}+\L_{sem}$) outperforms the modification with a ground truth semantic guidance ($\L_{FAWN}$ only). We attribute this to the joint training of semantic and geometry models. Predicted scene surfaces may deviate from ground truth during training, so using ground truth walls and floor as a guidance, we might target wrong normals. At the same time, \ours{} derives semantics and TSDF using the same backbone features, which makes these two outputs consistent.


\section{Conclusion}
\label{sec:conclusion}

We presented \ours, a modification of direct scene reconstruction methods, which considers the typical scene structure. Specifically, our modification determines walls and floor in a scene during training, and penalizes the corresponding surface normals for divergence from horizontal and vertical directions, respectively. This add-on not only eliminates surface malformations, such as pits and hills, but also closes the gaps in floor and walls. Implemented as a 3D sparse convolutional module, \ours{} can be incorporated into an arbitrary trainable pipeline that outputs TSDF, and be trained end-to-end on 3D semantic point clouds. During inference, 3D semantics is not needed, no extra assumptions are made, and no computational overhead is brought, so \ours{} does not limit usage scenarios. We showed that \ours{} improves over existing ways of considering 3D semantics, and is also beneficial for state-of-the-art TSDF reconstruction methods. 

\bibliographystyle{IEEEbib}
\bibliography{refs}

\end{document}